\documentclass[runningheads]{llncs}
\usepackage[linkcolor = blue]{hyperref}
\usepackage[T1]{fontenc}
\usepackage{multirow}
\usepackage{graphicx,verbatim}
\usepackage{color,soul}

\begin{document}
\title{Bridging Research and Practice: A Systematic Evaluation of Generalist and Dermatology-Specific Models in Clinical Skin Lesion Classification}
\titlerunning{Evaluating Generalist and Dermatology-Specific Models}
%

\author{
Emanoel dos Santos\inst{1}
\and
Kelvin Cunha\inst{1}
\and
Rodrigo Mota\inst{1}
\and
Fabio Papais\inst{1}
\and
Thales Bezerra\inst{1}
\and
Natalia Lopes\inst{1}
\and
Erico Medeiros\inst{1}
\and
Shirley Cruz\inst{2}
\and
Jessica Araujo\inst{2}
\and
Paulo Borba\inst{1}
\and
Tsang Ing Ren\inst{1}
}
\authorrunning{E. dos Santos et al.}
%
\institute{Centro de Informática, Universidade Federal de Pernambuco (UFPE), Brazil 
\email{\{etcs, kbc, raafm, flfp, tob2, nal, emm2, phmb, tir\}}@cin.ufpe.br \and
Hospital das Clinicas, Ebserh, UFPE, Pernambuco, Brazil}

\maketitle              
\begin{abstract}
The application of machine learning to dermatology has grown substantially in recent years, moving beyond proof-of-concept studies toward potential applications. However, clinical dermatology remains a challenging and still open problem. Diagnostic assessment is often ambiguous, and skin lesions exhibit high variability, compounded by differences in acquisition modality, device quality, and patient demographics. These factors hinder the development of robust models suitable for safe and equitable clinical use. To support translation into practice, it is essential to systematically evaluate how contemporary models generalize across heterogeneous data sources. In this work, we benchmark a diverse set of architectures on recent dermatology datasets, spanning dermoscopic images and smartphone-based clinical photographs. We assess the robustness of recent general-purpose and medical vision-language models, as well as foundation models, and compare them against task-specific dermatology classifiers, including embedding-based approaches and convolutional neural networks. Our study provides an evaluation of model performance under distribution shifts, modality changes, and demographic variability. By quantifying the gap between current state-of-the-art models and the requirements of clinical deployment, we aim to contribute to the development of reliable, accessible, and clinically applicable AI systems for dermatology.


\end{abstract}
\section{Introduction}

Skin cancer is among the most common malignancies worldwide, and timely recognition is essential on reducing patient risk of developing serious conditions~\cite{arnold2022global}. Clinical practice diagnosis relies on visual assessment through dermoscopy or standard photography, often under heterogeneous acquisition conditions~\cite{hasan2023survey}. Also, operator expertise and patient characteristics introduce substantial variability, making consistent and reliable assessment challenging~\cite{bezerra2025dermai,Pacheco2020-mn}. Any artificial intelligence system intended for clinical use must therefore demonstrate robustness across modalities and real-world data conditions. Recent advances in large-scale foundation models and vision-language models (VLMs) have generated considerable enthusiasm in medical imaging~\cite{sellergren2025medgemma,yuceyalcin2026hierarchical}. Their broad pretraining and multimodal capabilities suggest improved generalization and contextual reasoning compared to traditional task-specific convolutional networks. As a result, increasingly large models are being adopted for dermatology tasks, sometimes with limited domain-specific adaptation. However, high performance on curated benchmarks does not guarantee reliability across imaging modalities, datasets, or patient populations. Clinical translation requires systematic validation under realistic variability. Models must be evaluated across datasets collected under different protocols, and under distributions that reflect actual deployment scenarios. Without such evaluation, it is unclear whether newer large-scale architectures meaningfully outperform specialized dermatology models or whether they introduce new failures. In this work, we conduct a structured benchmark of diverse modeling paradigms for binary malignancy risk prediction across multiple dermatology datasets. We compare dermatology-specific embedding models, conventional convolutional networks, general representation learners, vision-language models, medical-adapted VLMs, and multimodal approaches incorporating metadata. Our evaluation spans modality-specific training, merged datasets, and cross-dataset testing. This study provides empirical evidence on the extent to which recent architectural advances translate into clinically meaningful robustness. Code is available at \url{https://github.com/TIC-13/derm-bench}.

\section{Methodology}

\subsection{Problem Formulation}

We address binary malignancy risk prediction from dermatological images. Given an input image $x$, the task is to predict a label $y \in \{0,1\}$, where $y=1$ denotes malignant and $y=0$ benign lesions. The objective is to evaluate model robustness across datasets and merged training configurations. Models were evaluated across multiple publicly available dermatology datasets encompassing dermoscopic and smartphone-based images. To assess robustness under heterogeneous conditions, we defined the following evaluation: (i) individual datasets,  (ii) merging datasets by modality (clinical or dermoscopic), (iii) and merging all datasets. For merged settings, datasets were combined at the image level while preserving original labels. To ensure fair comparison across model families identical train/test splits were used across models adopting the 70/15/15 split configuration, except where splits are already provided by the original source. Image preprocessing (resizing and normalization) was standardized and no test data leakage across datasets was allowed.

\subsection{Model Families}

We evaluated three model families reflecting current approaches in dermatological AI:

\begin{itemize}
    \item \textbf{Dermatology-Specific and General Embedding Models}: We evaluated domain-specific and general embedding backbones, including the derm-foundation~\cite{kolesnikov2020big,zhang2022contrastive} and the self-supervised DinoV3~\cite{simeoni2025dinov3}. For these models, the backbone was used as a frozen feature extractor and image embeddings were computed for all samples. A downstream classifier was trained on top of the embeddings (i.e, MLP, SVM, Random Forest, and XGBoost). This design decouples representation learning from classification and allows consistent comparison across different backbone models.
    \item \textbf{Convolutional and Vision Transformer Models}: We evaluated task-specific supervised architectures trained end-to-end, including ConvNeXt, EfficientNet, ResNet, and Vision Transformer (ViT). Models were fine-tuned on each dataset using cross-entropy loss.
    \item \textbf{Vision Language Models (VLMs)}: Large multimodal models were evaluated using prompt-based inference strategies, including: Dual-encoder configurations, simple prompts, structured prompts based on dermatological criteria, prompts augmented with patient metadata when available. We tested the models: BiomedCLIP~\cite{zhang2023biomedclip}, CLIP ViT-B/32, CLIP ViT-L/14~\cite{radford2021learning}, MedSigLIP-448~\cite{zhai2023sigmoid}, Gemma-3 27B, LLaMA-3.2-Vision 11B, LLaMA-3.2-Vision 90B, LLaMA-4 (16$\times$17B)~\cite{abdullah2025evolution}, Qwen-2.5VL 72B~\cite{bai2023qwen}.
\end{itemize}

\subsection{VLMs variations}

We evaluated multiple prompt and inference configurations to assess the behavior of large VLMs in dermatological malignancy prediction.

\begin{itemize}
    \item \textbf{Simple Prompt (Zero-Shot Classification)}: In the baseline configuration, the VLM receives the image and a direct instruction to classify the lesion as \textit{benign} or \textit{malignant}. The system prompt defines the model as an expert dermatologist, while the  user prompt restricts the output to a single-word label. This setup evaluates pure zero-shot visual-language reasoning without explicit clinical heuristics.
    \item \textbf{ABCDE-Guided Prompt}: To incorporate structured dermatological knowledge, we augmented the system prompt with an internal checklist based on the ABCDE criteria~\cite{duarte2021clinical} and the ``ugly duckling'' sign~\cite{gaudy2017ugly}. The model is instructed to apply these warning signs silently when making its decision, while still outputting only a single label. This configuration tests whether embedding domain heuristics into the prompt improves classification consistency and alignment with clinical reasoning.
    \item \textbf{Metadata-Augmented Prompt}: In this variant, structured clinical metadata (when available) is provided alongside the image. The system prompt explicitly instructs the model to integrate both visual and clinical information in its decision. This configuration evaluates whether contextual patient information enhances malignancy prediction.
    \item \textbf{Dual-Encoder Classification}: In the dual-encoder setup, the model processes image and label representations separately and predicts the most likely class via similarity matching.  Rather than generating free-text responses, the model selects between predefined class labels (\textit{benign}, \textit{malignant}). This approach evaluates discriminative multimodal alignment without reliance on generative prompting.
\end{itemize}

\subsection{Datasets}


We evaluated our models on publicly available dermatology datasets spanning dermoscopic and clinical imaging modalities. Dermoscopic benchmarks include HAM10000~\cite{tschandl2018ham10000},ISIC 2018~\cite{codella2019skin}, and ISIC 2024~\cite{kurtansky2024slice}. Clinical datasets include PAD-UFES-20 (PAD)~\cite{Pacheco2020-mn}, which provides smartphone images with structured patient metadata, HC~\cite{bezerra2025dermai}, reflecting heterogeneous real-world acquisition conditions, DDI~\cite{Daneshjou2022-cd}, designed to improve skin tone diversity and fairness evaluation, and SD-198~\cite{sun2016benchmark}, covering a broad range of skin diseases. To assess robustness under heterogeneous training conditions, we additionally constructed merged datasets combining sources: MERGED-ALL (all modalities), MERGED-CLINIC (clinical only), and MERGED-DERM (dermoscopic only), simulating multi-institutional deployment scenarios.

\subsection{Class Imbalance and metrics}

Given the inherent class imbalance in dermatology datasets, we evaluated three training strategies (i) Unbalanced training, (ii) Random undersampling, (iii) SMOTE-based oversampling. For embedding based pipelines, imbalance handling was applied during classifier training. For end-to-end CNN/ViT models, imbalance handling was implemented at the data sampling stage. The best-performing configuration per dataset was used for family-level comparison. Performance was assessed using standard metrics, as in the source datasets, we focus on the F1-Score to discuss our analyzes due to class imbalance and the clinical importance of balancing false positives and false negatives in malignancy risk prediction.


\section{Results}

\subsection{Overall Performance Across Datasets}

We first compared the best-performing configuration of each model family on each dataset. 
Table~\ref{tab:best_results} summarizes the highest F1 achieved per family and dataset. Embedding-based models achieved the highest average score ($88.6\%$), followed by CNN/ViT models ($86.6\%$), while VLMs achieved substantially lower performance ($65.8\%$). Thus, VLMs consistently ranked last across all evaluation settings. The performance gap between VLMs and the best vision-only approaches are substantial. In malignancy risk prediction, it may translate into a considerable increase in false negatives or false positives, directly affecting triage decisions. The persistence of this gap across dermoscopic, clinical, and merged settings suggests that model scale and multimodal pretraining alone do not ensure clinically reliable performance in dermatology.

\begin{table*}[t]
\centering
\caption{Best Macro F1-score per model family across datasets.}
\label{tab:best_results}
\begin{tabular}{lccc}
\hline
Dataset & VLM & CNN/ViT & Embeddings \\
\hline
PAD & 0.7671 & 0.9162 & 0.9343 \\
HC & 0.6123 & 0.8281 & 0.8760 \\
ISIC18 & 0.6026 & 0.8509 & 0.8738 \\
ISIC24 & 0.7048 & 0.8444 & 0.8333 \\
DDI & 0.6667 & 0.9112 & 0.8712 \\
HAM10000 & 0.5622 & 0.8859 & 0.9043 \\
SD-198 & 0.8426 & 0.8552 & 0.9638 \\
MERGED-ALL & 0.6114 & 0.8664 & 0.8783 \\
MERGED-CLINIC & 0.6375 & 0.8624 & 0.8888 \\
MERGED-DERM & 0.5794 & 0.8384 & 0.8503 \\
\hline
\textbf{Average F1} & \textbf{0.6587} & \textbf{0.8659} & \textbf{0.8874} \\
\hline
\end{tabular}
\end{table*}

When grouping datasets by modality, embedding and CNN/ViT models demonstrated stable performance across both dermoscopic and clinical images, supporting their robustness under varying acquisition conditions. In contrast, VLM performance showed higher variability and did not demonstrate consistent advantages on clinical photographs despite incorporating language-based reasoning. Notably, dermatology-specific embeddings achieved particularly strong performance on SD-198, indicating that domain-aligned representations can effectively capture clinically lesion patterns.

In the merged training configurations (MERGED-ALL, MERGED-CLINIC, MERGED-DERM), the relative ranking of model families remained unchanged. Embedding-based approaches maintained the highest average performance ($F1 \approx 88\%$), followed by CNN/ViT models ($F1 \approx 86\%$), while VLMs remained substantially lower ($F1 \approx 65\%$). This suggests that increasing dataset diversity alone does not compensate for architectural misalignment with the task, and that domain-specific visual representations remain critical for consistent malignancy assessment.

\subsection{Impact of Class Imbalance}

To assess the influence of imbalance mitigation strategies, we compared the best-performing embedding-based configuration under three training settings (unbalanced, random undersampling, and SMOTE). Table~\ref{tab:imbalance} reports the results obtained under each strategy. Overall, unbalanced training achieved the highest scores, followed by SMOTE and undersampling. While imbalance handling did not drastically alter performance in most datasets, certain settings exhibited notable sensitivity. For instance, undersampling improved performance on DDI ($87\%$ vs.\ $82\%$ unbalanced), whereas SMOTE led to performance degradation in ISIC24 and SD-198. These findings highlight that class rebalancing strategies must be carefully validated, as synthetic oversampling or aggressive undersampling may inadvertently affect sensitivity to malignant cases. Robust evaluation across imbalance settings is therefore essential before deployment in screening or triage scenarios.

\begin{table*}[t]
\centering
\caption{Best Macro F1-score achieved under different class imbalance strategies (embedding-based models).}
\label{tab:imbalance}
\begin{tabular}{lccc}
\hline
Dataset & Unbalanced & Undersampling & SMOTE \\
\hline
PAD & 0.9343 & 0.9124 & 0.9152 \\
HC & 0.8760 & 0.8680 & 0.8567 \\
ISIC18 & 0.8738 & 0.7797 & 0.8716 \\
ISIC24 & 0.8313 & 0.8333 & 0.7955 \\
DDI & 0.8224 & 0.8712 & 0.8071 \\
HAM10000 & 0.9022 & 0.8050 & 0.9043 \\
SD-198 & 0.9638 & 0.9638 & 0.9396 \\
MERGED-ALL & 0.8771 & 0.8583 & 0.8783 \\
MERGED-CLINIC & 0.8844 & 0.8872 & 0.8888 \\
MERGED-DERM & 0.8503 & 0.8069 & 0.8412 \\
\hline
\textbf{Average F1} & \textbf{0.8816} & \textbf{0.8586} & \textbf{0.8698} \\
\hline
\end{tabular}
\end{table*}

\subsection{Architectures validation}

To provide a more granular view of performance, we summarize results at the  architecture level rather than by model family. We report the average of the best F1-scores obtained per dataset, grouped into three clinically relevant categories: dermoscopic datasets (ISIC18, ISIC24, HAM10000, MERGED-DERM), clinical datasets (PAD, HC, DDI, SD-198, MERGED-CLINIC), and fully merged datasets (MERGED-ALL).

\begin{table*}[t]
\centering
\caption{Average Macro Best F1-score per architecture across evaluation categories.}
\label{tab:arch_summary}
\begin{tabular}{lccc}
\hline
Architecture & Dermoscopic & Clinical & Merged-All \\
\hline
Derm-Foundation & 0.8308 & 0.8872 & 0.8714 \\
DINOv3 ViT-7B/16 & 0.8503 & 0.8888 & 0.8783 \\
DINOv2 Giant & 0.8383 & 0.8479 & 0.8641 \\
ConvNeXt (best) & 0.8373 & 0.8624 & 0.8619 \\
EfficientNet (best) & 0.8171 & 0.8218 & - \\
Best VLM (prompt-based) & 0.5794 & 0.6723 & 0.6114 \\
\hline
\end{tabular}
\end{table*}

Across dermoscopic datasets, large vision backbones (EfficientNet, DINOv3, Derm-Foundation) achieved comparable performance. In clinical datasets, Derm-Foundation and EfficientNet showed the strongest robustness, while VLMs remained substantially lower. In the merged-all setting, embedding-based architectures maintained stable performance, whereas VLMs did not exhibit improvement despite increased dataset diversity. These results indicate that performance differences are primarily driven by representation quality rather than model scale alone. Architectures trained or aligned with dermatological visual features demonstrate greater stability across acquisition modalities. Notably, DINO-based architectures achieved performance comparable to dermatology-specific trained models, despite relying on general self-supervised feature representations. Fine-tuning strategies further improved their results, highlighting the strength of transferable visual representations when properly adapted to the task.

\subsection{VLMs Variations}

For this test, we report results only on datasets for which structured patient metadata is available (HC, PAD, ISIC18, ISIC24), enabling a fair comparison across prompt-based and metadata-augmented VLM configurations.

\begin{table*}[t]
\centering
\caption{F1-score comparison of VLM variants on datasets with available patient metadata.}
\label{tab:vlm_variations}
\resizebox{0.8\textwidth}{!}{
\begin{tabular}{llcccc}
\hline
Setting & Model & HC & PAD & ISIC18 & ISIC24 \\
\hline
\multirow{4}{*}{Dual Encoder}
 & BiomedCLIP & 0.3362 & 0.3951 & 0.3375 & 0.3041 \\
 & CLIP-B/32 & 0.5882 & 0.4866 & 0.5185 & 0.5382 \\
 & CLIP-L/14 & 0.3191 & 0.4866 & 0.1740 & 0.6116 \\
 & MedSigLIP & 0.5686 & 0.6387 & 0.4815 & 0.4755 \\
\hline
\multirow{5}{*}{Prompt + ABCDE}
 & Gemma-3 27B & 0.5313 & 0.5504 & 0.2040 & 0.7048 \\
 & LLaMA-3.2V 11B & 0.3087 & 0.4090 & 0.1367 & 0.2823 \\
 & LLaMA-3.2V 90B & 0.4962 & 0.5217 & 0.3725 & 0.5980 \\
 & LLaMA-4 16$\times$17B & 0.6123 & 0.6476 & 0.1632 & 0.4494 \\
 & Qwen-2.5VL 72B & 0.5175 & 0.6213 & 0.4243 & 0.6088 \\
\hline
\multirow{5}{*}{Prompt + Metadata}
 & Gemma-3 27B & 0.4693 & 0.7381 & 0.3203 & 0.6127 \\
 & LLaMA-3.2V 11B & 0.4756 & 0.4290 & 0.2010 & 0.2823 \\
 & LLaMA-3.2V 90B & 0.3003 & 0.4398 & 0.2174 & 0.3650 \\
 & LLaMA-4 16$\times$17B & 0.3099 & 0.4141 & 0.2482 & 0.4168 \\
 & Qwen-2.5VL 72B & 0.4693 & 0.7671 & 0.6026 & 0.5507 \\
\hline
\multirow{5}{*}{Prompt (Simple)}
 & Gemma-3 27B & 0.4930 & 0.5685 & 0.3716 & 0.6451 \\
 & LLaMA-3.2V 11B & 0.5357 & 0.6835 & 0.5524 & 0.4744 \\
 & LLaMA-3.2V 90B & 0.4356 & 0.3662 & 0.4197 & 0.4457 \\
 & LLaMA-4 16$\times$17B & 0.3454 & 0.3483 & 0.3093 & 0.3967 \\
 & Qwen-2.5VL 72B & 0.4872 & 0.5035 & 0.5680 & 0.5644 \\
\hline
\end{tabular}}
\end{table*}

Dual-encoder models exhibit more stable behavior than prompt-based VLMs, suggesting that discriminative image-text alignment is less sensitive to prompt design than generative reasoning. Among them, MedSigLIP and CLIP-B/32 consistently outperform BiomedCLIP, despite the latter being trained on biomedical data. Prompt engineering using ABCDE heuristics leads to moderate gains in selected cases (e.g., ISIC24), but improvements are inconsistent and highly model-dependent. Incorporating patient metadata yields the strongest improvements for a subset of models, particularly on PAD, indicating that contextual clinical information can partially compensate for visual ambiguity. 

Overall, these results suggest that current VLMs lack the robustness required for clinical deployment. Performance appears constrained less by prompt sophistication or model scale and more by the intrinsic difficulty of prediction from images alone. Simply applying general-purpose VLMs in zero-shot settings, or augmenting them with additional contextual information, is insufficient to achieve performance comparable to specialized models. Rather than reflecting an inherent limitation of VLMs, these findings highlight the importance of task-specific adaptation and proper training. Although general VLMs may appear to solve the target task, deploying them without proper domain-specific adaptation can pose substantial risks to application reliability.

\subsection{Qualitative Analysis}

\begin{figure*}[t]
    \centering
    \includegraphics[width=\textwidth]{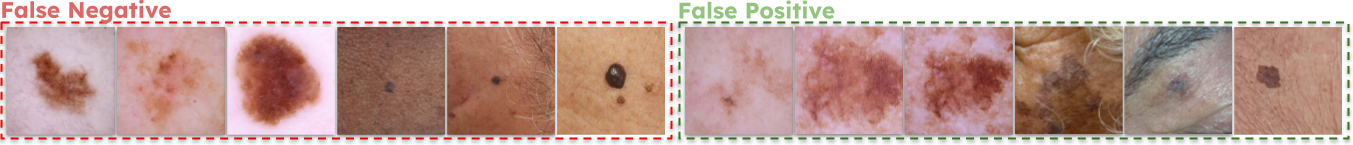}
    \caption{Qualitative false negative (left) and false positive (right) cases in malignancy prediction on dermoscopic (HAM10000) and clinical (HC) data.}
    \label{fig:fp_fn}
\end{figure*}

Figure~\ref{fig:fp_fn} illustrates representative false positive and negative cases. Several benign lesions exhibit irregular borders, color heterogeneity, and partial asymmetry, while some malignant cases present relatively symmetric shapes and limited color variation. This visual overlap highlights a fundamental limitation of image-only malignancy prediction: morphological features alone may be insufficient to reliably separate benign from malignant lesions in all cases. In clinical practice, dermatological assessment extends beyond visual inspection. Patient age, lesion location, temporal evolution, symptoms, and prior history contribute substantially to risk estimation. The absence of such contextual information may explain persistent errors even in high-capacity models. Purely visual classification, particularly under binary malignancy labels, imposes intrinsically ambiguous decision boundaries that cannot be fully resolved through model scale alone.

\section{Conclusion}

We presented an evaluation of contemporary architectures for binary malignancy risk prediction across heterogeneous dermatology datasets. Our analysis shows that dermatology-aligned embedding models and strong vision backbones consistently outperform large vision-language models. Increasing architectural scale or incorporating multimodal pretraining did not translate into improved clinical robustness. Instead, architectures with domain-aligned visual representations demonstrated the most stable performance under dataset variability. These findings emphasize that rigorous cross-dataset validation and task-specific representation learning remain essential prerequisites for reliable clinical deployment of AI systems in dermatology.

\begin{credits}
\subsubsection{\ackname} The project was supported by the Ministry of Science, Technology, and Innovation of Brazil, with resources from Law No. 8,248, dated October 23, 1991, under the scope of the PPI-SOFTEX, coordinated by Softex and published under RESIDÊNCIA EM TIC 63 – ROBÓTICA E IA – FASE II, DOU 23076.043130/2025-27. For partially supporting this work, we would like to thank INES.IA (National Institute of Science and Technology for Software Engineering Based on and for Artificial Intelligence) \url{www.ines.org.br}, CNPq grant 408817/2024-0.
\end{credits}

\section*{Disclosure of Interests}
The authors declare that they have no competing interests.

%
%
%
\bibliographystyle{splncs04}
\bibliography{paper-5997}
%




\end{document}